\pdfoutput=1

\documentclass[11pt]{article}

\usepackage{acl}

\usepackage{times}
\usepackage{latexsym}

\usepackage[T1]{fontenc}

\usepackage[utf8]{inputenc}
\DeclareUnicodeCharacter{0327}{\c{c}}
\DeclareUnicodeCharacter{0301}{\'{ }}

\usepackage{microtype}
\usepackage{mathtools}

\begin{document}

\title{GATE: A Challenge Set for Gender-Ambiguous Translation Examples}

\author{Spencer Rarrick \And Ranjita Naik  \And Varun Mathur \And Sundar Poudel  \And     Vishal Chowdhary
\thanks{All authors are affiliated with Microsoft.}
\thanks{Contact author at \texttt{\scriptsize vishalc@microsoft.com}.}}
\maketitle

\begin{abstract}
Although recent years have brought significant progress in improving translation of unambiguously gendered sentences, translation of ambiguously gendered input remains relatively unexplored. When source gender is ambiguous, machine translation models typically default to stereotypical gender roles, perpetuating harmful bias. Recent work has led to the development of "gender rewriters" that generate alternative gender translations on such ambiguous inputs, but such systems are plagued by poor linguistic coverage. To encourage better performance on this task we present and release GATE, a linguistically diverse corpus of gender-ambiguous source sentences along with multiple alternative target language translations. We also provide tools for evaluation and system analysis when using GATE and use them to evaluate our translation rewriter system.

\end{abstract}

\section{Introduction}

Gender is expressed differently across different languages. For example, in English the word \emph{lawyer} could refer to either a male or female individual, but in Spanish, \emph{{\color{teal}abogada}} and \emph{{\color{purple}abogado}} would be used to refer to a female or a male lawyer respectively. This frequently leads to situations where in order to produce a single translation, a translator or machine translation (MT) model tends to choose an arbitrary gender to assign to an animate entity in translation output where it was not implied by the source. In this paper, we refer to this phenomenon as \emph{arbitrary gender marking} and to such entities as Arbitrarily Gender-Marked Entities (AGMEs).

Translation with arbitrary gender marking is a significant issue in MT because these arbitrary gender assignments often align with stereotypes, perpetuating harmful societal bias (\citealp{stanovsky-etal-2019-evaluating}; \citealp{chloe-mt-bias}). For example, MT models will commonly translate the following (from English to Spanish):

\begin{center}
    \emph{The surgeon $\xRightarrow{\text{MT}}$ {\color{purple}El cirujano} (m)}
    
    \emph{The nurse $\xRightarrow{\text{MT}}$ {\color{teal}La enfermera} (f)}
\end{center}

Progress has been made to remedy this using a "gender rewriter" -- a system that transforms a single translation with some set of gender assignments for AGMEs into a complete set of translations that covers all valid sets of gender assignments for a source sentence into the target language \citep{DBLP:journals/corr/abs-1809-02208}. Using a rewriter:

\begin{center}
    \emph{The surgeon}
    
    \vspace{.25cm}
    $\big\Downarrow$ MT
    \vspace{.25cm}
    
    \emph{{\color{purple}El cirujano}} (m) 
    
    \vspace{.25cm}
    \hspace{.55cm}$\big\Downarrow$ rewriter
    \vspace{.25cm}

    \emph{{\color{teal}La cirujana}} (f)
    
    \emph{{\color{purple}El cirujano}} (m)
\end{center}

Although a step in the right direction, these rewriters often have poor linguistic coverage and only work correctly in simpler cases. Google Translate has publicly released such a system for a subset of supported languages, and we observe two error cases\footnote{as observed on Mar 6, 2023}:

\begin{enumerate}
    \item It does not rewrite when necessary: \emph{The director was astonished by the response of the community.} produces only one translation corresponding to masculine director.  
    \item It rewrites partially, or incorrectly: \emph{I'd rather be a nurse than a lawyer} produces two translations but only lawyer is reinflected for gender (nurse is feminine in both).
\end{enumerate}

To facilitate improvement in coverage and accuracy of such rewriters and reduce bias in translation, we release GATE\footnote{Data and evaluation code available at https://github.com/MicrosoftTranslator/GATE}, a test corpus containing gender-ambiguous translation examples from English (en) into three Romance languages (\citep{Vincent-1988}): Spanish (es), French (fr) and Italian (it). Each English source sentence\footnote{A few non-sentence utterances are also included as well, such as noun-phrases and sentence fragments} is accompanied by one target language translation for each possible combination of masculine and feminine gender assignments of AGMEs \footnote{The majority of source sentences contain only one AGME and thus two translations}:

\begin{center}
\emph{I know \textbf{a Turk} who lives in Paris.}

$\big\Downarrow$ it

\emph{Conosco {\color{teal}una turca} che vive a Parigi.} (f)

\emph{Conosco {\color{purple}un turco} che vive a Parigi.} (m)    
\end{center}

GATE is constructed to be challenging, morphologically rich and linguistically diverse. It has $\sim 2000$ translation examples for each target language, and each example is annotated with linguistic properties  (coreferent entities, parts of speech, etc.). We additionally propose a set of metrics to use when evaluating gender rewriting modules.

This corpus was developed with the help of bilingual linguists with significant translation experience for each of our target languages (henceforth \emph{linguists}). Each is a native speaker in their respective target language. We spoke in depth with our linguists about the nuances of gender-related phenomena in our focus languages and we share our analysis of the relevant aspects and how they impact our work and the task of gender rewriting.

Along with the corpus, we also provide tools for evaluation and system analysis when using GATE and use them to evaluate our own translation rewriter system.

\section{GATE Corpus}

We present GATE corpus, a collection of bilingual translation examples designed to challenge source-aware gender-rewriters. The linguists were asked to compile roughly 2,000 examples for each target language, with the hope that this would be sufficient for good variety along several dimensions: sentence lengths, sentence structures, vocabulary diversity, and variety of AGME counts.

\subsection{Anatomy of an Example}
\label{section:Anatomy}

Each example in the data set consists of an English sentence with at least one AGME, and a set of alternative translations into the given target language corresponding to each possible combination of male/female gender choices for each AGME. Variation among the alternative translations is restricted to the minimal changes necessary to naturally and correctly indicate the respective gender-markings.

We also mark several category features on each example, such as what class of animate noun AGMEs belong to (profession, relationship, etc), what grammatical role they play in the sentence (subject, direct object, etc), sentence type (question, imperative, etc) and several other phenomena. These are discussed in more detail in Appendix \ref{appendix:categories}, as well as statistics over each language's corpus.

Additionally, each example is accompanied by a list of AGMEs as they appear in the English source, as well as their respective masculine and feminine translations found in the translated sentences. For multi-word phrases, we asked annotators to enclose the head noun in square brackets. For example, if \emph{police officer} is translated to \emph{policía} in Spanish, the English field could include \emph{police [officer]}. 

The same entity may be referred to multiple times in the same sentence through coreference. We asked annotators to indicate coreferent mentions of AGMEs are by joining them with '='. For example, in the following \emph{en-es} example, the English AGME field would contain "nurse=lawyer".

\begin{center}
\emph{I'd rather be a \textbf{nurse} than a \textbf{lawyer}.}

$\big\Downarrow$ es

\emph{Prefiero ser {\color{teal}enfermera} que {\color{teal}abogada}.} (f)

\emph{Prefiero ser {\color{purple}enfermero} que {\color{purple}abogado}.} (m)    
\end{center}

Finally, In cases where an AGME is represented by a pronoun that is elided in the translation, it will be represented by the nominative case form and be enclosed in parentheses. For example, in the following example, the Spanish AGME field would contain \emph{(yo)}: 

\vspace{.1cm}
\hspace{2.cm} \emph{I am \textbf{tired}.}

\hspace{3cm}  $\big\Downarrow$ es

\hspace{1.5cm} \emph{Estoy {\color{teal}cansada}.} (f)

\hspace{1.5cm} \emph{Estoy {\color{purple}cansado}.} (m)
\vspace{.25cm}

\begin{table*}
\centering
\begin{tabular}{|l|llll|l|}
\hline
\textbf{Data Set} & \textbf{< 10} & \textbf{10-19} & \textbf{20-29} & \textbf{>= 30} & \textbf{Total}\\
\hline   
Spanish 1 AGME	 & 477 & 722 & 197 & 105 & 1,501  \\
Spanish 2+ AGMEs  & 70  & 176 & 56  & 21  & 323   \\
\hline				                             
French 1 AGME	& 704 & 661 & 171 & 14	& 1,550  \\
French 2+ AGMEs  & 177 & 222 & 41  & 4   & 444   \\
\hline					                             
Italian 1 AGME	 & 397 & 867 & 195 & 48	 & 1,507  \\
Italian 2+ AGMEs & 93  & 500 & 139 & 30	 & 762   \\
\hline          
\end{tabular}
\caption{\label{sentence lengths}
Distribution of lengths (words) of English utterance per target language and AGME count}
\end{table*}

\subsection{Arbitrarily Gender-Marked Entities}
In this paper, we use \emph{animate entity} (or just \emph{entity}) to refer to an individual or group for which a referential gender could be implied in either the source or target language\footnote{For simplicity, we limit our discussion of gender and linguistics to masculine and feminine within the scope of this paper, but we do not intend to imply that gender is limited in this way.}. Usually this will refer to humans, but may also be extended to some animals and mythical or sentient beings. For example, \emph{cat} is generally translated into Spanish as \emph{{\color{purple}gato}}, but \emph{{\color{teal}gata}} is also frequently used to refer to a female cat. Following \citet{dahl_2000}, we use \emph{referential gender} to refer to an entity's gender as a concept outside of any linguistic considerations. 

To qualify as an AGME, an entity's referential gender must be ambiguous in the source sentence, but implied by one or more words in the target translation. Compared to Romance languages, there are relatively few ways that gender is denoted through word-choice in English. Most notably, English uses a handful of gendered pronouns and possessive adjectives (\emph{{\color{teal}she}},  \emph{{\color{teal}her}}, \emph{{\color{teal}hers}}, \emph{{\color{purple}he}}, \emph{{\color{purple}him}}, \emph{{\color{purple}his}}), as well as a relatively small number of animate nouns that imply a gender (e.g. \emph{{\color{teal}mother}}, \emph{{\color{purple}father}}, \emph{{\color{teal}masseuse}}, \emph{{\color{purple}masseur}}, etc). There is also often a correlation between certain proper names and referential gender (e.g. \emph{Sarah} is traditionally a female name and \emph{Matthew} is traditionally male), but we do not consider this a reliable enough signal for gender determination unless they are a well known public figure (e.g. \emph{{\color{purple}Barrack Obama}} is known to be male). We follow \citet{vanmassenhove2021genderit} in this.

Additionally, an AGME must have some gender marking in the translation. In the following English-Italian example,

\vspace{.1cm}
\emph{{\color{olive}I} heard {\color{purple}the thief} insult \textbf{his  interlocutor}.}

\hspace{3cm}  $\big\Downarrow$ \emph{it}

\hspace{-.5cm}\emph{{\color{olive}Io} ho sentito {\color{purple}il ladro} insultare {\color{teal}la sua interlocutrice}.}

\hspace{-.5cm}\emph{{\color{olive}Io} ho sentito {\color{purple}il ladro} insultare {\color{purple}il suo interlocutore}.}

\vspace{.1cm}

\emph{\textbf{interlocutor}}$\rightarrow$\emph{{\color{teal}interlocutrice}} (f) / \emph{{\color{purple}interlocutore}} (m) is an AGME, while \emph{{\color{purple}thief}}$\rightarrow$\emph{{\color{purple}ladro}} and \emph{{\color{olive}I}}$\rightarrow$\emph{{\color{olive}Io}} are not. \emph{Thief} is unambiguously male because of its coreference with \emph{his} in the source, while \emph{I} has ambiguous gender which is not marked in the target.

\subsection{Corpus Development Process}

The linguists were asked to aim for a distribution of sentences lengths ranging from very short (< 10 words) to complex (> 30) words. Actual example counts are shown in Table \ref{sentence lengths}.
Of the 2,000 examples for each language, linguists were asked to include roughly the following breakdown:
\vspace{-.1pc}
\begin{itemize}
    \setlength\itemsep{-0.4em}
    \item 1,000 single animate noun AGME
    \item 500   single pronoun AGME
    \item 500   with two or more AGMEs
\end{itemize}
\vspace{-.3pc}

Linguists were given details of the various categories and attributes listed in section \ref{appendix:categories} and asked to find sentences such that each such category is well represented (depending on the relative ease of finding such sentences). Linguists were also asked to prioritize diversity of animate nouns where possible. They were allowed to pull examples sentences from natural text or construct them from scratch as they saw fit. However, except for a small number of toy examples, we asked that they include only sentences that were natural in both English and their target language, and could reasonably appear in some imaginable context. 

We provided samples of web-scraped data that had been filtered with various heuristics to help identify sentences fitting some of the harder-to-satisfy criteria. For example, we used Stanza \citep{qi2020stanza} to filter some web-scraped data for those containing an animate noun marked as an indirect object and provided this to the linguists. In some cases these sentences were used directly, and in others they were modified slightly to fit the requirements.

Throughout the process, we prioritized diversity of sentence structure, domain and vocabulary. Rather than produce a representative sample, our intention was to produce a corpus that would challenge any tested systems on a wide range of phenomena. 

\section {Evaluation with GATE}

\subsection{Gender Rewriting}

Our goal in developing this corpus is to facilitate the generation of multiple translations covering all valid gender assignments. One strategy for producing such a set of translations is to first use an MT model to produce a default translation and then use a rewriter to generate one or more alternative translations with other gender assignments \citep{DBLP:journals/corr/abs-1809-02208}.
$$\text{source} \xRightarrow{\text{MT}} \text{translation} \xRightarrow{\text{rewriter}} \text{\{all translations\}}$$

\subsection{Evaluation Methodology}

We formalize the task of gender rewriting on a single-AGME sentence as follows: given the source sentence $src$, target translations corresponding to male and female referent entities, and a rewrite direction (M to F or F to M), produce an output target translation with the alternative gender from the original translation. We will refer to the original input translation as $tgt_0$, the desired/reference translation as $tgt_1$ and the output generated by the rewriter as $hyp$:
$$rewriter(src, tgt_0) = hyp \sim tgt_1$$

For this task, we consider looking at exact full-sentence matches between $hyp$ and $tgt_1$ to be the most sensible approach for evaluation. We do not give partial credit for changing the gender markings on only a subset of the words to those found in $tgt_1$. Doing so will generally result in a sentence that is either grammatically incorrect due to newly introduced agreement errors, or for which the semantics has changed in an unacceptable way, such as a changed coreference. Because of this, we find sentence-similarity measures such as BLEU \citep{papineni-etal-2002-bleu} and words error rate not to reflective of a user's experience.

The rewriter may also produce a null output, meaning that only the default translation will be produced. This is necessary because in real-world scenarios, many sentences will not contain AGMEs. When AGMEs are present, it may still be preferable to produce null output over a low confidence rewrite if accuracy errors are judged to be more costly than coverage errors. 

We calculate precision as the proportion of correct alternatives among those attempted, i.e. that were non-null outputs. Because there are no true negatives in GATE, recall can be calculated as the proportion of correct alternatives produced among all sentences, including null outputs. Using these definitions of precision and recall, we also find $F_0.5$ to be a useful overall metric, prioritizing precision while still incorporating coverage.

While we have focused our discussion of evaluation on sentences containing a single AGME, which typically should produce exactly two alternative translations, GATE also includes a smaller number of examples with more than one AGME. These have more than two alternative translations and thus more than one correct output for a rewriter. We do not formalize evaluation on this subset here but believe that the data set will be useful in evaluating rewriting systems capable of producing multiple outputs for multiple sets of gender assignments.

\subsection {System Overview}

We use GATE to evaluate our translation gender rewriter, which follows a pipeline approach, roughly similar to \citet{habash-etal-2019-automatic}. 

The system receives as input the original source sentence ($src$) and a default translation ($tgt_0$) with the specified language pair. The following components are then applied: 

\textbf{AGME Identifier} – We first attempt to find AGMEs in the sentence pair to determine whether rewriting is appropriate. We leverage an AllenNLP coreference model to detect ambiguously gendered entities in the source sentence \citep{Lee2018HigherorderCR}. We use a dependency parse generated by Stanza \citep{qi2020stanza} and a gendered vocab list to identify gender-marked animate entities in the target sentence. 

\textbf{Candidate Generator} - For each word position in $tgt_0$, we use a lookup table to find all possible alternate gender variants for the word in that position. We compose the word-level variant sets to build a set of sentence-level hypotheses, while applying grammatical constraints to prune incoherent hypotheses. This yields a set of candidate rewrites.

\textbf{Translation Scorer} - Finally, we use a Marian translation model \citep{mariannmt} to score each rewrite candidate as a translation of source sentence. If no candidates have scores close to the $tgt_0$, We return a null output. Otherwise we choose . 

\subsection{Experimental Results}

We evaluate our system for rewriting quality on GATE in both masculine-to-feminine and feminine-to-masculine directions. To simulate runtime efficiency constraints, we impose a cutoff of 20 maximum source words. Any input sentence longer than this is treated as a null output and therefore a false negative.  

\begin{table}[!ht]
    \centering
    \begin{tabular}{|l|c||l|l|l|}
    \hline
        Language & Direction & P & R & F0.5  \\ \hline\hline
        Spanish & F$\rightarrow$M & 0.97 & 0.50 & 0.82   \\ \hline
        Spanish & M$\rightarrow$F & 0.95 & 0.40 & 0.74 \\ \hline
        French & F$\rightarrow$M & 0.97 & 0.28 & 0.65 \\ \hline
        French & M$\rightarrow$F & 0.91 & 0.27 & 0.61 \\ \hline
        Italian & F$\rightarrow$M & 0.96 & 0.47 & 0.79 \\ \hline
        Italian & M$\rightarrow$F & 0.91 & 0.32 & 0.67 \\ \hline
    \end{tabular}
    \caption{\label{system stats} Our rewriter's scores on GATE for each target language and rewrite direction}
\end{table}

From these results we can see that our system performs best for Spanish in both directions, and in the female-to-male direction across all language pairs. Both trends can be explained to an extent by the properties of the translation models. High quality training data for English-Spanish is more plentiful than for the other two languages, leading to a higher quality model in general. As noted earlier, translation models have been shown to skew towards stereotypical gender assignments, which are more heavily weighted towards masculine forms. Therefore, it is not too surprising that when rewriting in this direction, the translation model is more likely to prefer an incorrect rewrite candidate.

\subsection{End-to-End Evaluation}

In our envisioned scenario, a gender rewriter would operate on the output of an MT system. It is unlikely, however, that direct MT output will consistently match GATE's translations word-for-word. As a result, references cannot be directly utilized, and human annotation is required to assess the output of a rewriter alongside machine translation (MT) or any integrated system that generates a series of gender alternative translations from a single source sentence. One consideration is that a parallel sentence from GATE may no longer contain an AGME when machine translated, as the MT output may be unmarked for gender.

In order to test our combined system end-to-end, we sampled 200 source sentences from GATE and used our production MT models to translate them into Spanish, and then pass that output to our rewriter. We then ask annotators to examine the source sentence and all translation outputs, and to provide the following annotations:

\vspace{-.1pc}
\begin{itemize}
    \setlength\itemsep{-0.4em}
    \item If two translations are produced, mark true positive if the following are true (otherwise false positive):
    \begin{itemize}
        \item Is the target gender-marked for an ambiguous source entity?
        \item Were all the words marking gender of AGME changed correctly?
        \item Were only the words marking gender of AGME changed?
    \end{itemize}
    
    \item If only one translation is produced, is the target gender marked for an ambiguous source entity? 
    
    \item If there are multiple AGMEs:
    \begin{itemize} 
        \item If two valid translations are produced mark as a true positive.
        \item If only one translation is produced mark as a true negative.
        \item Otherwise mark as a false positive.
    \end{itemize}
\end{itemize}
\vspace{-.3pc}

We also retrieve translations for these sentences from an online English-Spanish translation system that can produce masculine and feminine alternative translations for this translation direction. We asked annotators to annotate these translations in the same manner. 

\begin{figure}[t]
    \centering
       \includegraphics[width=9cm]{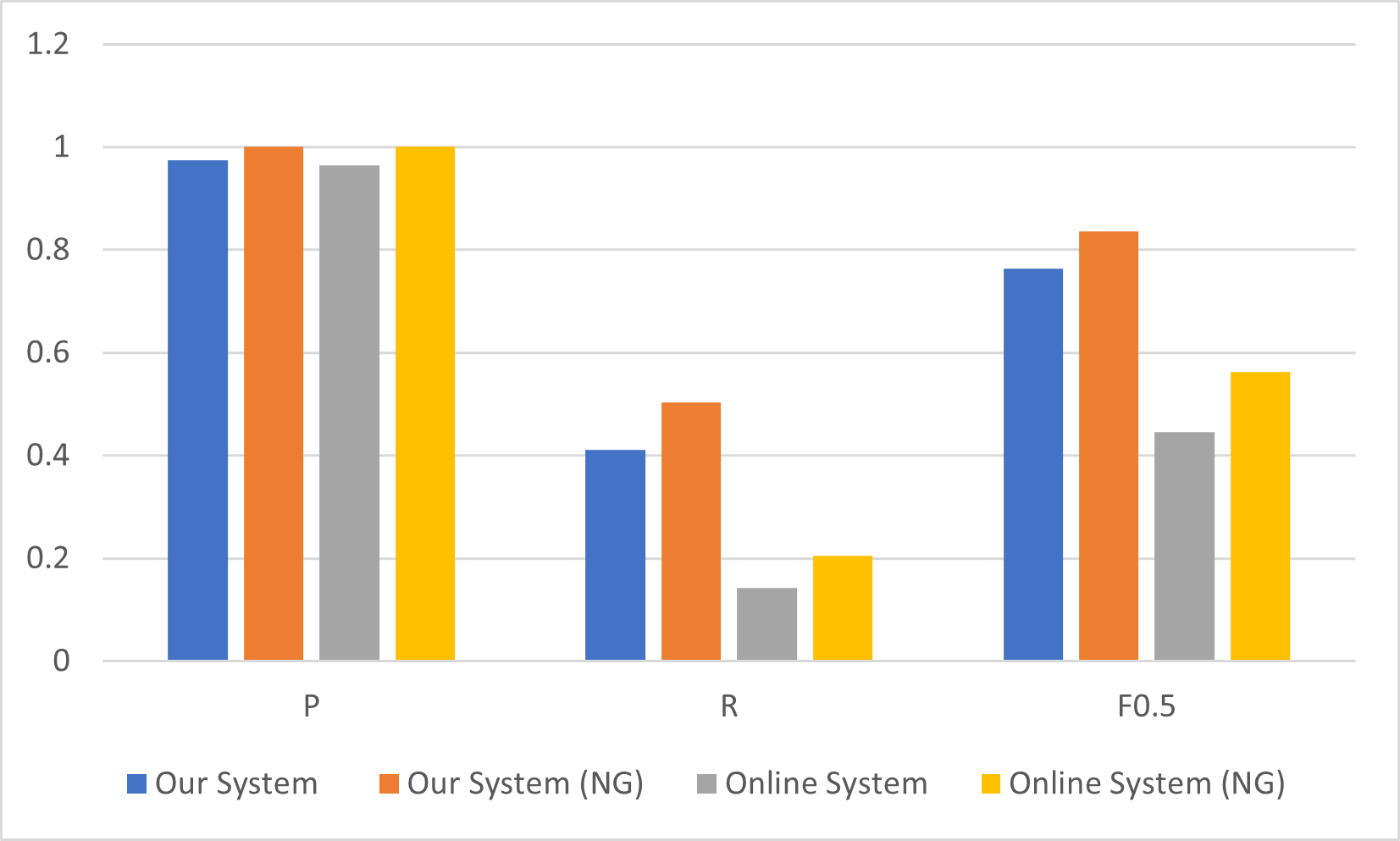}
      \caption{End-to-end scores for our system and an online translation system.}
      \label{endToEndEvalPlot}
  \end{figure}

Finally, we also asked annotators to mark source sentences for which the speaker is reasonably likely to know the referent's gender, and therefore use of a masculine generic should be less likely (see \ref{masculinegeneric}). We evaluate quality on that subset as well for each system, in rows marked NG ( \emph{(non-generic)}). Results are presented in Table \ref{endToEndEval} and visualized in Figure \ref{endToEndEvalPlot}.

\begin{table}[!ht]
    \centering
    \begin{tabular}{|l|l|l|l|}
    \hline
        & P & R & F0.5  \\ \hline
        Our System & 0.97 & 0.41 & 0.76 \\ \hline
        Our System (NG) & 1.00 & 0.50 & 0.84 \\ \hline \hline
        Online system & 0.96 & 0.14 & 0.45 \\ \hline
        Online system (NG) & 1.00 & 0.21 & 0.56 \\ \hline
    \end{tabular}
    \caption{\label{endToEndEval} end-to-end scores for our system and an online translation system. NG rows are calculated only on \emph{non-generic} sentences}
\end{table}

Both systems heavily favor precision over recall, and recall is somewhat higher on the \emph{non-generic} portion of the data. Overall, our system demonstrates significantly better coverage.

A full, end-to-end evaluation should include testing both sentences with and without AGMEs. As each instance in GATE involves at least one AGME, we suggest enhancing GATE with instances from \citet{renduchintala2021} and \citet{vanmassenhove2021genderit}, which feature unequivocally gendered source entities. In future work, we intend to develop a supplemental data set for GATE containing various types of negative examples: unambiguous source entities, entities that are unmarked in both source and target, and inanimate objects whose surface forms are distractors (e.g. depending on context, \emph{player} and \emph{cleaner} may refer to either objects or people).

\section{Linguistic Background}


\subsection{Gender in Romance Languages}

In Spanish, French and Italian, all nouns have a grammatical gender -- either masculine or feminine. For inanimate objects, this gender is fixed and often arbitrary; for example, in French, \emph{{\color{teal}chaise}} (chair) is feminine, while \emph{{\color{purple}canapé}} (couch) is masculine. When a noun or pronoun refers to an animate entity, its grammatical gender will, with some notable exceptions, match the referential gender of that entity. \citep{Vincent-1988}

In these languages, referential gender of entities is frequently marked through morphology of an animate noun (e.g. \emph{en-es}: \emph{lawyer} $\Rightarrow$ \emph{{\color{teal}abogada}}(f), \emph{{\color{purple}abogado}}(m)) or through agreement with gendered determiners, adjectives and verb forms. 

\subsection{Dual Gender and Epicene}
\label{dual-gender}

Some animate nouns are \emph{dual gender}, meaning that the same surface form is used for both masculine and feminine, such as French \emph{artiste} (artist) (\citealt{corbett_1991} as cited in \citealt{genderacrosslanguages-vol2}). However, other clues to the artist's gender may exist in a French sentence through gender agreement with other associated words. For example, \emph{The tall artist} could be translated into French as \emph{{\color{teal}La grande artiste}}(f) or \emph{{\color{purple}Le grand artiste}}(m). Here, grammatical gender of translations of \emph{the} (\emph{{\color{teal}la}} (f) / \emph{{\color{purple}le}} (m)) and \emph{tall} (\emph{{\color{teal}grande}} (f), \emph{{\color{purple}grand}} (m)) must match the referential gender of the referent noun.

Dual-gender determiners and adjectives exist as well, such as Spanish \emph{mi} (my) and \emph{importante} (important). So, for example, Spanish \emph{mi huésped importante} (My important guest) has no gender marking. Similarly, in French and Italian, some determiners may contract before vowels to lose their gender marking. Feminine and masculine forms of \emph{the} in French, \emph{{\color{purple}le}} and \emph{{\color{teal}la}}, both contract before vowels (and sometimes \emph{h}) to become \emph{l'}, so \emph{l'artiste} (the artist) is not marked for gender.

While typically an entity's referential gender will align with its grammatical gender, these languages each contain a handfull of \emph{epicene} nouns. These are nouns whose grammatical gender is fixed, regardless of the referential gender of the referent (\citealt{lebonusage} as cited in \citealt{genderacrosslanguages-vol3}). Most notable among these is the direct translation of \emph{person} into each of the target languages, which is always grammatically feminine: \emph{{\color{teal}La persona}} (\emph{es,it}) or \emph{{\color{teal}La personne}} (\emph{fr}). We also find some language-specific epicene nouns. For example, these Italian words are always grammatically feminine: \emph{{\color{teal}la guardia}} (guard), \emph{{\color{teal}la vedetta}} (sentry), \emph{{\color{teal}la sentinella}} (sentry), \emph{{\color{teal}la recluta}} (recruit), \emph{{\color{teal}la spia}} (spy).  \footnote{Color-coding in this paragraph corresponds only to grammatical gender, while referential gender is ambiguous in these expressions.}

\subsection{Pronouns}

Similarly to English, some pronouns in Romance languages are inherently gendered, while others are not. Entities referred to by gender-neutral pronouns, such as Spanish \emph{yo} (I) and \emph{tú} (you) commonly become gender-marked through predicative gender-inflecting adjectives. Further complicating these cases, subject pronouns are frequently omitted in Spanish and Italian (but notably not in French) as the subject can be inferred from verb morphology (\citealt{genderacrosslanguages-vol2}, pp. 189, 252). This means that in some cases, the AGME in a sentence pair may be a zero-pronoun, such as English \emph{I am \textbf{tired}} being translated to Spanish as \emph{estoy {\color{teal}cansada}} (f) or \emph{estoy {\color{purple}cansado}} (m). There is no overt subject in these translations corresponding to \emph{I}, but the subject is implied by the verb form \emph{estoy}.

\subsection{Coreference}

Another common pattern is that of coreferent mentions of a single entity, which must by definition have the same referential gender, and usually but not always the same grammatical gender. For example, in the following sentence, \emph{friend} and \emph{nurse} are the same individual and we would typically expect them to share the same referential gender in a direct translation into any of the target languages. 

    \vspace{.25cm}
    \hspace{1.5cm}\emph{My best friend is a nurse}
    \vspace{.25cm}

In cases where one coreferent mention is an epicene noun as described in \ref{dual-gender}, the grammatical genders of those mentions may in fact differ. In the following sentence, the described individual is unambiguously male. The phrase \emph{{\color{teal}una buena persona}} (a good person) is grammatically female, while \emph{{\color{purple}un mal amigo}} (a bad friend) and \emph{{\color{purple}él}} (he) are grammatically male. \footnote{In this example, color-coding indicates grammatical gender of each mention as it appears the Spanish translation}

\vspace{.25cm}
\hspace{.5cm}\emph{{\color{purple}He} is {\color{teal}a good person} but {\color{purple}a bad friend}.}

\hspace{3cm}  $\big\Downarrow$ es

\hspace{.5cm}\emph{{\color{purple}Él} es {\color{teal}una buena persona}, pero {\color{purple}un mal amigo}.}
\vspace{.1cm}

\subsection{Masculine Generics}
\label{masculinegeneric}

Traditionally, many languages, including Spanish, French and Italian, employ a paradigm known as masculine generics. Under this paradigm, feminine forms are considered to be explicitly gender-marked, while masculine forms should be used in situations where referential gender is unclear. Specifically, when referential gender is unknown by the speaker, or a mixed-gender group is known to contain at least one male individual, defaulting to grammatically masculine forms is generally considered correct in the language standard\footnote{In recent years there is some explorations of using novel, gender-neutral forms in these contexts}. In this sense, masculine gender marking does not imply the exclusion of female-identifying individuals, but a feminine gender marking would imply the exclusion of male-identifying individuals. (\citealt{genderacrosslanguages-vol2}, \citeyear{genderacrosslanguages-vol3})

In most cases where a masculine generic might be used, we nonetheless ask our linguists to provide an alternative translation with feminine gender-marking. Language critics have noted that the use of masculine generics can evoke an association with 'male' (\citealt{genderacrosslanguages-vol3}, pp. 101), and so we believe that inclusion of a feminine generic variant fits our mission of promoting inclusive language use. Our linguists were asked to annotate such generic mentions with the label INDF (\emph{indefinite gender}), so that users who wish to follow a stricter interpretation can exclude these examples in their evaluations. However, upon analysis of our corpus we noted that  this annotation was only consistently applied to the Italian data.

\section{Related Work}
A slew of challenge sets has been proposed for evaluating gender bias in Machine Translation. 

\textbf{MuST-SHE} (\citet{bentivogli-etal-2020-gender} ; \citet{savoldi-etal-2022-morphosyntactic}
comprises approximately 1000 triplets consisting of audio, transcript, and reference translations for en-es, en-fr, and en-it languages. Each triplet is classified based on the gender of the speaker or explicit gender markers, such as pronouns, as either masculine or feminine. Furthermore, the dataset contains an alternative incorrect reference translation for every correct reference translation that alters the gender-marked words.

\textbf{WinoMT} \citet{stanovsky-etal-2019-evaluating} is a challenge set that comprises English sentences containing two animate nouns, one of which is coreferent with a gendered pronoun. Based on the context provided in the sentence, a human can easily identify which animate noun is coreferent and thus deduce the gender of the person described by that noun. By evaluating the frequency with which an MT system generates a translation with the correct gender for that animate noun, one can measure the extent to which the system depends on gender stereotypes rather than relevant context.

\textbf{SimpleGEN} \citet{renduchintala-etal-2021-gender} on the English-Spanish (en-es) and English-German (en-de) language pairs. It includes a test set consisting of short sentences with straightforward syntactic structures. Each source sentence includes an occupation noun and a clear indication of the gender of the person described by that noun. In other words, the source sentence provides all the necessary information for a model to generate occupation nouns with the correct gender.

\textbf{The Translated Wikipedia Biographies}\footnote{https://ai.googleblog.com/2021/06/a-dataset-for-studying-gender-bias-in.html} dataset comprises 138 documents containing human translations of Wikipedia biographies from English to Spanish and German. Each document comprises 8-15 sentences, providing a context for gender disambiguation evaluation across sentences.

\textbf{MT-GenEval} \citet{currey-etal-2022-mt} is a dataset that includes gender-balanced, counterfactual data in eight language pairs. The dataset ensures that the gender of individuals is unambiguous in the input segment, and it comprises multi-sentence segments that necessitate inter-sentential gender agreement.

Regarding the work on addressing ambiguously gendered inputs, \citet{habash-etal-2019-automatic} tackle translation of ambiguous input by treating it as a gender classification and reinflection task when translating English into Arabic. Their approach focuses on the first-person singular cases. Given a gender-ambiguous source sentence and its translation, their system generates an alternative translation using the opposite gender. Additionally, they create a parallel corpus of first-person singular Arabic sentences that are annotated with gender information and reinflected accordingly. \citet{alhafni2021arabic} expand on the work of \citet{habash-etal-2019-automatic} by adding second person targets to the Arabic Parallel Gender Corpus, as well as increasing the total number of sentences. 

Google Translate announced\footnote{https://ai.googleblog.com/2020/04/a-scalable-approach-to-reducing-gender.html} an effort to address gender bias for ambiguously gendered inputs by showing both feminine and masculine translations. They support this feature for English to Spanish translation, as well as several gender-neutral languages into English.

Regarding debiasing in the monolingual context, \cite{zmigrod-etal-2019-counterfactual} propose a generative model capable of converting sentences inflected in masculine form to those inflected in feminine form, and vice versa, in four morphologically rich languages. Their work focuses on animate nouns.

In terms of rewriting text in English, \citet{vanmassenhove-etal-2021-neutral} and \citet{https://doi.org/10.48550/arxiv.2102.06788} propose rule-based and neural rewriting models, respectively, that are capable of generating gender-neutral sentences.

\section{Conclusion}
 
We have presented GATE, a corpus of hand-curated test cases designed to challenge gender rewriters on a wide range of vocabulary, sentence structures and gender-related phenomena. Additionally, we provide an in-depth analysis of many of the nuances of grammatical gender in Romance languages and how it relates to translation. We also suggest metrics for gender rewriting and provide tools to aid with their calculation. Through this work we aim to improve the quality of MT output in cases of ambiguous source gender, as well as facilitate the development of better and more inclusive natural language processing (NLP) tools in general.

We look forward to future work in improving GATE and related projects. We aim to add additional languages pairs to GATE and investigate translation directions into English. We also hope to supplement with additional data, including negative examples. Finally, we plan to explore use of gender-neutral language use in various languages and how it can be incorporated into NLP applications.

\section{Bias Statement}

In this work, we propose a test set to evaluate translation of ambiguously gendered source sentences by NMT systems. Our work only deals with English as the source and is currently scoped to Romance languages as the target. To construct our test set, we have worked with bilingual linguists for each target language. We plan to increase scope of both source and target languages in future work.

Through this work, we hope to encourage and facilitate more inclusive use of natural language processing technology, particularly in terms of gender representation. In recent years, there is significant ongoing movement in the way gender manifests in languages use. One form that this takes is in new gender-neutral language constructs in Romance languages such as French, Spanish and Italian to accommodate gender underspecificity and non-binary gender identities. We support the development of this more representative and inclusive language, and endeavor to find ways to support it through technology. In this work, however, for the sake of simplicity we restrict our scope to language as used to express gender along more conventionally binary lines, and we therefore do not consider non-binary language or word forms. We are working with both language experts and non-binary-identifying individuals to expand the scope to include non-binary and gender-underspecified language in future work.

\bibliography{anthology,custom}
\bibliographystyle{acl_natbib}

\appendix
\section{Category Labels}
\label{appendix:categories}

There are a wide range of linguistic phenomena that can interact with gender in translation. We have devised several category labels that can be applied to examples. In order to promote diversity within the corpus, linguists were asked to ensure that a certain minimum number of examples are included for each such label. This also has the benefit of helping pinpoint weaknesses in an evaluated system. For example, a rewriting system may perform well when the ambiguous noun is the subject of the sentence, but do poorly when it is a direct object. We hope to include per-category evaluation and analaysis of our system in a future version of this work.

Unless otherwise stated, category labels are determined based on the target sentence set rather than the source sentence, as this is generally the more important input to the rewriter. A single example will typically have multiple labels.

\begin{itemize}
    \item \textbf{Grammatical Role categories:} An AGME is a subject (SUBJ), direct object (DOBJ) indirect object (IOBJ), subject complement (SCMP), or object of a preposition (OPRP, excluding indirect objects) 
    \item \textbf{Animate Noun categories:} profession (PROF, e.g. doctor), Religion (REL, e.g. Bhuddist), Nationality (NAT, e.g. Italian), Family and other relationships (REL, e.g. neighbor), Non-Human (NHUM, e.g. cat, vampire), Other (OTH, e.g. winner, accused)
    \item \textbf{Adjectives and past participles:} attributive (AATR), predicative (APRD), past-participle form as an adjective (PPA), past-participle form not as an adjective (PPNA), Adjective modifies non-ambiguous noun (ANAN). Most of these distinctions are included to test a rewriter's ability to distinguish between adjective surface forms that should be modified along with key nouns and those that should not.
    \item \textbf{Sentence Types categories:} Headline (HEAD), sentence fragment (FRAG), question (QUES), imperative (IMPR), Ambiguous noun in a subordinate clause (SUBC)
    \item \textbf{Other categories:} Plural ambiguous noun (PLUR), indefinite i.e. does not refer to an entity concretely known by the speaker, e.g. "Where can I find a good doctor?" (INDF), Requires agreement across different clauses with noun that was ambiguous in source (DFCL), Distinct animate nouns behave as a single group and are \emph{gender-linked} (GLNK)
\end{itemize}

\begin{table*}
\centering
\begin{tabular}{|l|lll|p{0.6\linewidth}|}
\hline
\textbf{Label} & \textbf{es} & \textbf{fr} & \textbf{it} & \textbf{description} \\
\hline
\multicolumn{5}{|l|}{\textbf{Semantic Type}} \\
\hline
PROF & 1168 & 490 & 1208 & Profession word  \\
NAT  & 118  & 249 & 157  & Nationality or locality membership   \\
REL  & 25   & 150 & 29   & Religious affiliation \\
FAM  & 327  & 250 & 192  & Family or other relationship \\
NHUM & 2    & 40  & --   & Non-Human \\
OTH  & 580  & 941 & 708  & Other \\
\hline
\multicolumn{5}{|l|}{\textbf{Grammatical role}} \\
\hline
SUBJ & 1638 & 1221 & 1573 & Subject \\
SCMP & 118  & 185  & 121  & Subject complement \\
DOBJ & 181  & 328  & 399  & Direct object \\
IOBJ & 136  & 275  & 165  & Indirect object \\
OPRP & 250  & 279  & 518  & Object of preposition \\
VOC  & 3    & --   & 4    & Vocative \\
POSC & 80   & --   & 289  & Possessive complement \\
\hline
\multicolumn{5}{|l|}{\textbf{Sentence Type}} \\
\hline
QUES & 124  & --   & --   & Question \\
FRAG & 49   & 101  & --   & Sentence Fragment \\
IMPR & 14   & 135  & --   & Imperative \\
\hline
\multicolumn{5}{|l|}{\textbf{Adjective-related}} \\
\hline
APRD & 82   & 359  & 213  & Predicative adjective agreeing with AGME\\
AATR & 293  & 190  & 315  & Attributive adjective agreeing with AGME\\
ANAN & 97   & 1026 & --   & Adjective modifying a word other than AGME\\
PPA  & 361  & 172  & 290  & Adjective has same surface form as a past participle\\
APPS & --   & 35   & 22   & post-positive adjective -- remove??\\
\hline
\multicolumn{5}{|l|}{\textbf{Pronoun subtype}} \\
\hline
PERS & --   & 219  & 146  & Personal pronoun \\
RELA & --   & 15   & 13   & Relative pronoun \\
DEMO & --   & 64   & 28   & Demonstrative pronoun \\
POSS & 80   & --   & --   & Possesive pronoun \\
DROP & 157  & --   & --   & AGME is a dropped/zero pronoun \\
IPRO & --   & 369  & 53   & Indefinite pronoun \\
\hline
\multicolumn{5}{|l|}{\textbf{Other}} \\
\hline
PLUR & 991 & 1110 & 1042 & Plural \\
INDF & --  & --   & 229  & Indefinite/masculine generic could apply\\
DFCL & 136 & 113  & --   & Changed words in alternatives cross clause boundaries \\
GLNK & --  & 94   & --   & "gender-link" -- AGMEs are not coreferent but conceptually linked, different genders would be unnatural \\

\hline          
\end{tabular}
\caption{\label{Cateogry Labels}
Counts of sentences with each category label per language. '--' indicates that this language was not annotated for this label}
\end{table*} 

\end{document}